\documentclass[sigconf]{acmart}
\usepackage{placeins}
\AtBeginDocument{%
  }
    
\setcopyright{acmlicensed}
\copyrightyear{2025}
\acmYear{2025}
\acmDOI{XXXXXXX.XXXXXXX}

\acmConference[ACM SIGSPATIAL '25]{4rd ACM SIGSPATIAL International Workshop on
Searching and Mining Large Collections of Geospatial Data (GeoSearch’25)}{November 3--6, 2025}{Minneapolis, MN}

\acmISBN{978-1-4503-XXXX-X/2018/06}

\begin{document}
\title{Detecting Legend Items on Historical Maps Using GPT-4o with In-Context Learning}

\author{Sofia Kirsanova, Yao-Yi Chiang}
\email{{kirsa002, yaoyi}@umn.edu}
\affiliation{%
  \institution{University of Minnesota}
  \city{Minneapolis}
  \state{MN}
  \country{USA}
}

\author{Weiwei Duan}
\affiliation{%
  \institution{Inferlink Corporation}
  \city{El Segundo}
  \state{CA}
  \country{USA}
}
\email{wduan@inferlink.com}

\begin{abstract}
  Historical map legends are critical for interpreting cartographic symbols. However, their inconsistent layouts and unstructured formats make automatic extraction challenging. Prior work focuses primarily on segmentation or general optical character recognition (OCR), with few methods effectively matching legend symbols to their corresponding descriptions in a structured manner. We present a method that combines LayoutLMv3 for layout detection with GPT-4o using in-context learning to detect and link legend items and their descriptions via bounding-box predictions. Our experiments show that GPT-4 with structured JSON prompts outperforms the baseline, achieving 88\% F-1 and 85\% IoU, and reveal how prompt design, example counts, and layout alignment affect performance. This approach supports scalable, layout-aware legend parsing and improves the indexing and searchability of historical maps across various visual styles. 
\end{abstract}

\ccsdesc[500]{Information systems~Geographic information systems}
\ccsdesc[300]{Information systems~Information retrieval}

\keywords{Map Digitization, Geospatial information retrieval, Map Layout Analysis, In-Context Learning}
  

\maketitle

\section{Introduction}

Historical geological maps contain a large amount of information about rock units, mineral deposits, and other features. To utilize these maps effectively, we must also understand their legends, which explain the meaning of symbols using their colors and patterns. Without the legend, a map is just visual marks on paper; with the legend, it becomes structured information that can be searched, compared, and analyzed. This makes understanding map legends an important step for turning scanned maps into digital, searchable data.

At the same time, working with map legends is a challenging task. Different map series and time periods use very different layouts. Some legends place items in neat lists, while others have irregular spacing, custom shapes, or even multiple columns. Additionally, there is no large annotated dataset of map legends, making it challenging to train deep learning models in a supervised manner.

These challenges are important in the broader context of geospatial search and mining. Many libraries, archives, and research groups are scanning thousands of historical maps; however, without automated legend interpretation, these collections remain difficult to search. A system that can detect and structure legend information would enable downstream tools to connect map symbols with their corresponding geological meanings.

By converting legend area information to machine-readable form, this work directly supports the goals of GeoSearch. Each detected legend item with its description act as ``metadata'' that can be used for retrieval and search across map collections. For instance, maps can be queried by geological unit names, symbols, or color patterns once their legends are digitized. In this way, legend interpretation becomes a foundation for scalable, multimodal geospatial search and mining across historical map archives.

This work is part of the DIGMAPPER\cite{digmapper} project, a modular system for automated map digitization developed under the DARPA CriticalMAAS program. DIGMAPPER introduced layout analysis, feature extraction, and georeferencing modules for scanned geological maps. In this paper, we expand the legend detection module in DIGMAPPER. Our method combines a layout-aware vision model to crop legend areas with a large language model GPT-4o used in an in-context learning (ICL) mode. By providing GPT-4o with a few structured examples, we guide it to predict bounding boxes for legend items and their linked descriptions on new maps. We evaluate the method on a curated set of U.S. Geological Survey (USGS) maps.

\noindent\textbf{Contributions.} 
This paper extends DIGMAPPER by focusing on the legend-detection module and contributes the following:
\begin{itemize}
    \item A training-free, in-context learning (ICL) approach using GPT-4o to detect and link legend items and descriptions on historical maps.
    \item A structured JSON prompting format that produces machine-readable bounding boxes, which helps to integrate with geospatial search and digitization pipelines.
    \item An empirical analysis of prompt design, including the effect of varying the number of in-context examples (5, 10, 15, and 20) to identify the optimal configuration.
\end{itemize}

\FloatBarrier 
\section{Related Work}

Historical map digitization has been studied for many years, with work ranging from feature extraction \cite{chiang2009automatic,arteaga2013historical} to automatic georeferencing \cite{luft2021automatic,gede2021automatic, howe2019deformable}. These efforts have shown that traditional computer vision and machine learning methods can extract specific objects, such as roads, intersections, or building footprints \cite{budig2015active,heitzler2020cartographic}. However, most of these techniques focus on the map content itself, while the legend area, which defines the semantic meaning of symbols, is often ignored.

Recent progress in document layout analysis has introduced models such as LayoutLMv3 \cite{layoutlmv3} and LayoutParser \cite{layoutparser}, which combine text, image, and layout features to understand structured documents. These methods show promise for tasks where spatial arrangement and multimodal input matter, such as detecting tables, figures, or complex page layouts. In the geospatial domain, similar approaches have been adapted to historical maps, for example, by segmenting map sheets into content and legend regions \cite{wu2023cross,lin2024hyper}. 

Very few studies explicitly address the challenge of extracting legend items. Some prior work has examined symbol recognition on topographic maps \cite{miao2017symbols,huang2023pointsymbol}, or linking text and features in historical map corpora \cite{li2020automatic,kim2023mapkurator}. More recently, LLMs have been proposed for geospatial tasks \cite{li2023geolm}, but their application to legends is still new. ICL \cite{incontext_survey} allows LLMs like GPT-4 \cite{gpt4o} to adapt to diverse layouts without retraining, which is particularly useful given the variety of cartographic styles. Our work builds on these advances by framing legend interpretation as a few-shot ICL task, using structured prompts that pair example items with descriptions.

Legends act as a “semantic key” that makes scanned maps searchable by attributes such as rock type or fault symbols. By automatically detecting and linking legend items to descriptions, our method contributes to scalable geospatial retrieval and mining and complements spatial indexing, content-based retrieval, and multimodal search \cite{li2020automatic}.

\FloatBarrier 
\section{Method}
\label{sec:method}

\noindent\textbf{Overview.}
Our approach builds on recent progress in map layout analysis and LLMs for structured reasoning. The goal is to automatically detect pairs of legend symbols (visual items) and their corresponding descriptions (text regions) from scanned geological maps. To achieve this, we design a pipeline with two main steps: (1) isolate the legend area from the full map, and (2) use GPT-4o in an ICL setting to generate bounding boxes for legend–description pairs.

\subsection{Legend Area Segmentation}
The first step is to segment each map into content and legend regions. We adopt a layout-aware vision model, specifically a fine-tuned LayoutLMv3, LARA,\footnote{\url{https://github.com/DARPA-CRITICALMAAS/uncharted-ta1/blob/main/pipelines/segmentation/README.md}}, which is applied to classify regions and identify the block that contains the legend. This step is important because historical maps vary widely: some have legends on the right side, others on the bottom, and a few use irregular or multi-column legend structures. After segmentation, we crop the legend area as an image file.

\subsection{In-Context Learning with GPT-4o}
The second step extracts bounding boxes for the legend items and their associated descriptions. We frame this as an ICL task: GPT-4o is guided to predict outputs based on structured examples.

The prompt given to GPT-4o has three parts:
\begin{itemize}
    \item \textbf{Example legend image:} a cropped legend area from an annotated map.
    \item \textbf{JSON prompt:} a structured text block containing the task definition, and a set of annotated pairs (legend item + description) from the example image, represented as bounding box coordinates.
    \item \textbf{Target legend image:} the cropped legend area of a new (unseen) map.
\end{itemize}

The JSON prompt specifies the task, lists 15 pairs of examples from the annotated legend, and includes a placeholder entry for the target map with coordinates set to ''??''. An example is shown below:

\begin{verbatim}
{
  "task": "Given a scanned map legend area, detect legend 
  items and their descriptions coordinates",
  "examples from example_map_legend.tiff": [
    {
      "legend_item": [6630.85, 472.34, 6779.79, 560.64],
      "description": [6214.89, 572.34, 7186.17, 621.28]
    },
    {
      "legend_item": [4985.96, 1233.62, 5145.11, 1324.26],
      "description": [4572.34, 1244.68, 5342.55, 1298.94]
    }
    // ... 13 more pairs
  ],
  "predictions for target_map_legend.tiff": [
    {
      "legend_item": ["??", "??", "??", "??"],
      "description": ["??", "??", "??", "??"]
    }
  ]
}
\end{verbatim}

GPT-4o responds with bounding box coordinates for the detected pairs in the target legend. The output is a JSON object, replacing the placeholders with numeric coordinates:

\begin{verbatim}
{
  "predictions for target_map_legend.tiff": [
    {
      "legend_item": [4700.25, 678.40, 4852.67, 750.95],
      "description": [4302.14, 690.50, 5100.80, 738.80]
    },
    {
      "legend_item": [3150.10, 910.25, 3305.20, 980.30],
      "description": [2800.00, 920.40, 3500.90, 965.00]
    }
    // ... additional detected pairs
  ]
}
\end{verbatim}


\begin{figure}[!t]
    \centering
    \includegraphics[width=\linewidth]{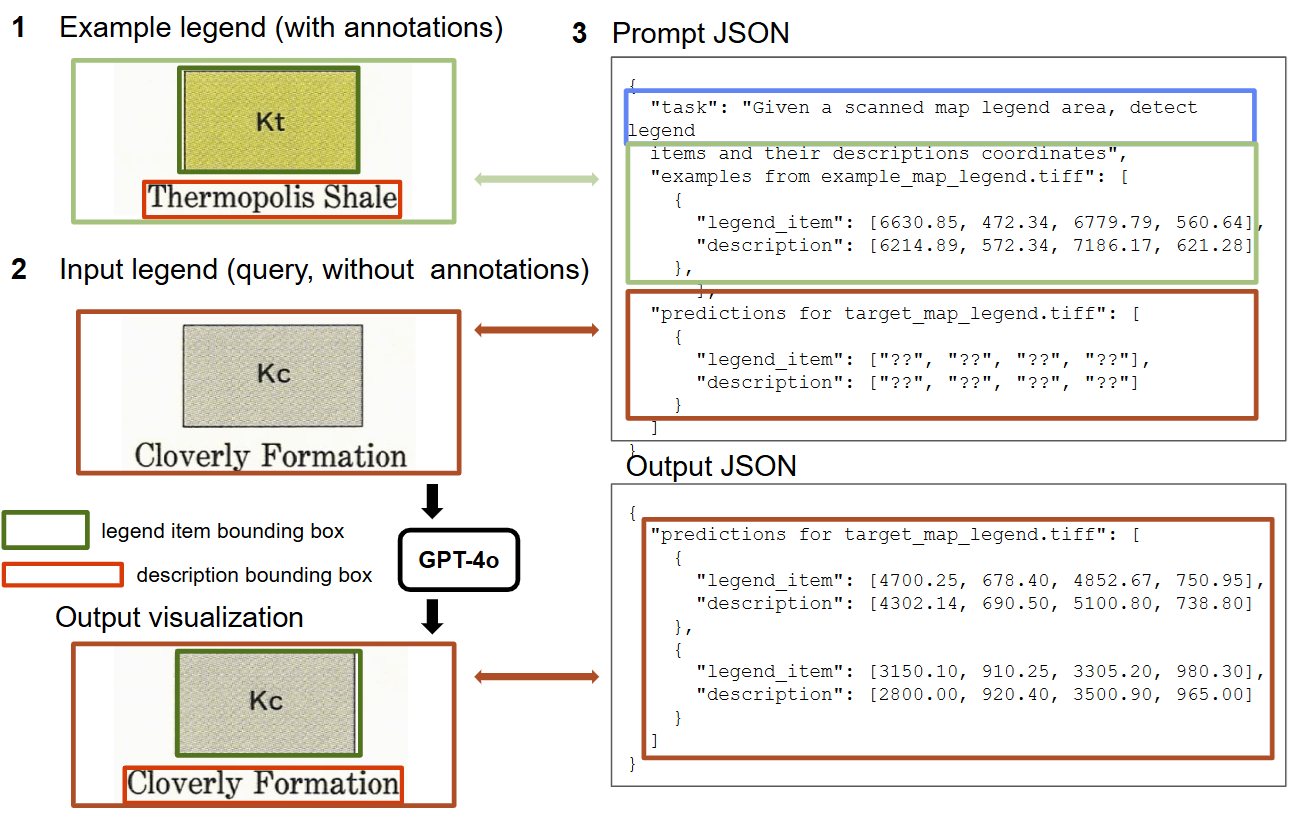}
    \caption{Workflow of our approach for extracting legend–description pairs with GPT-4o. The inputs are (1) an annotated example legend area, (2) a target legend area, and (3) a JSON-formatted prompt. The outputs are bounding boxes coordinates for legend–description pairs in the legend area.}
    \label{fig:legend_description_workflow}
\vspace{-5pt}
\end{figure}

\noindent\textbf{Relevance to Geospatial Search.}
The structured legend outputs generated by our approach provide semantic metadata that directly improves geospatial search. Each detected item-description pair links a visual symbol to its corresponding geological meaning, forming a searchable index for map features. This enables queries such as finding all maps that contain a specific rock unit or fault symbol in scanned collections. When integrated into larger pipelines, these outputs enable multimodal search across text, color, and spatial patterns. Color information from each legend item can be stored (for example, the dominant fill color), which allows users to search for maps that contain similar color patterns such as specific rock-unit shades.

\section{Experiments}
\label{sec:experiments}

\noindent\textbf{Dataset.}
We evaluated our method on 40 annotated maps randomly selected from the DARPA–USGS historical map dataset\cite{darpa_usgs_2025_dataset}. Each map is a high-resolution \texttt{.tiff} scan with ground-truth annotations for legend areas, legend items, and their corresponding descriptions. These annotations serve as the basis for computing the detection accuracy.

\noindent\textbf{Evaluation Metrics.}
We report two standard object detection metrics: Intersection over Union (IoU) and F1 score. A predicted bounding box is considered correct if its IoU with a ground truth box is greater than or equal to 0.5. From these matches, we compute precision, recall, and F1. IoU measures geometric overlap, while F1 balances the rates of false positives and false negatives.

\noindent\textbf{Effect of In-Context Examples.}
Because our approach relies on GPT-4o in an ICL setting, the number of examples provided in the prompt is an important factor. We tested prompts with 5, 10, 15, and 20 example pairs drawn from annotated legends. 

As shown in Table~\ref{tab:icl_examples}, performance improves steadily from 5 to 15 examples. With only five examples, the model achieves reasonable results but often fails to capture complex layouts. At 10 examples, both IoU and F1 scores increase by several points. The best results are obtained with 15 examples, which provide the most consistent predictions across different legend layouts. When 20 examples are included, performance slightly declines; overly long prompts introduce noise and distract the model from the core task.

\begin{table}[htbp]
\centering
\caption{Effect of number of in-context examples on legend item and description detection (IoU and F1 at threshold 0.5). Best scores are highlighted in bold.}
\label{tab:icl_examples}
\begin{tabular}{l|cc|cc}
\toprule
\textbf{\# Examples} & \multicolumn{2}{c|}{\textbf{Legend Item}} & \multicolumn{2}{c}{\textbf{Description}} \\
                     & \textbf{IoU} & \textbf{F1} & \textbf{IoU} & \textbf{F1} \\
\midrule
5   & 0.78 & 0.82 & 0.76 & 0.84 \\
10  & 0.82 & 0.86 & 0.81 & 0.88 \\
15  & \textbf{0.85} & \textbf{0.88} & \textbf{0.84} & \textbf{0.92} \\
20  & 0.83 & 0.86 & 0.82 & 0.88 \\
\bottomrule
\end{tabular}
\end{table}

\begin{figure}[!t]
    \centering
    \includegraphics[width=\linewidth]{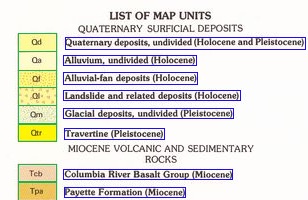}
    \caption{Example of GPT-4o predictions on a USGS legend (15 in-context examples). Bounding boxes show detected item–description pairs}
    \label{fig:legend_description_workflow}
\vspace{-5pt}
\end{figure}

\noindent\textbf{Overall Results.}
Table~\ref{tab:overall_results} summarizes the overall segmentation and detection results across all 40 test maps. LARA is used to crop the legend area, followed by GPT-4o for item–description detection. Compared with the baseline of LayoutLMv3 alone for the whole task, the GPT-based approach for the item-description detection step achieves the highest scores.

\noindent\textbf{Implementation Details.}
All experiments used the GPT-4o model (April 2025 version) accessed through the OpenAI API. Each query has one example legend image, one target legend image, and a JSON prompt averaging 1.2k input tokens and 0.6k output tokens.The inference pipeline was deterministic, with fixed prompts and temperature = 0.

\begin{table}[!t]
\centering
\caption{Comparison of different methods for legend detection.}
\label{tab:overall_results}
\begin{tabular}{l|cc|cc}
\toprule
\textbf{Method} & \multicolumn{2}{c|}{\textbf{Legend Item}} & \multicolumn{2}{c}{\textbf{Description}} \\
                & \textbf{IoU} & \textbf{F1} & \textbf{IoU} & \textbf{F1} \\
\midrule
LayoutLMv3              & 0.69 & 0.72 & 0.75 & 0.79 \\
\textbf{GPT-4o (15 ex.)}   & \textbf{0.85} & \textbf{0.88} & \textbf{0.84} & \textbf{0.92} \\
\bottomrule
\end{tabular}
\end{table}

\begin{figure}[!t]
    \centering
    \includegraphics[width=\linewidth]{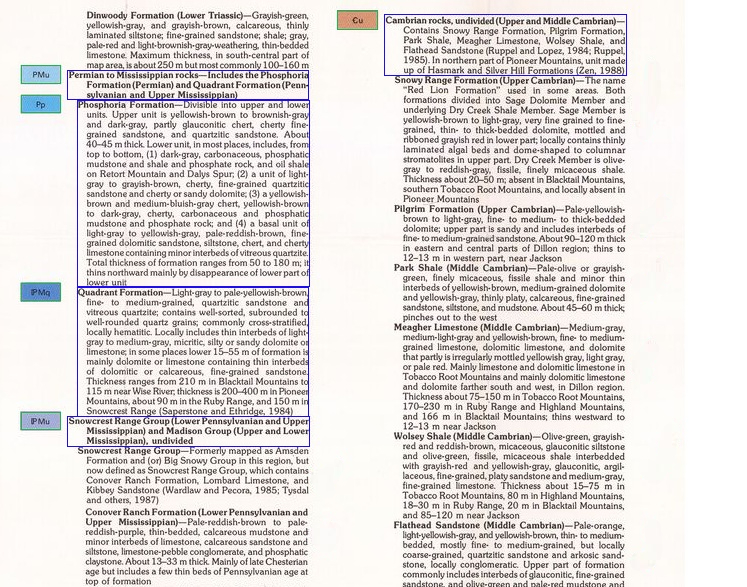}
    \caption{Failure case of legend item–description detection. The multi-column layout led to oversized bounding boxes and several missed pairs.}
    \label{fig:legend_failure}
\vspace{-5pt}
\end{figure}

\noindent\textbf{Discussion.}
These results show that GPT-4o with ICL effectively extracts structured legend item–description pairs from historical maps. Performance improves with more examples in the prompt, suggesting that providing additional context helps the model generalize to diverse layouts. The best results are achieved with 15 examples, which strike a balance between sufficient guidance and avoiding overly long prompts.

Figure~\ref{fig:legend_failure} shows a failure case from a map with a dense multi-column legend. Several descriptions were missed or merged into oversized bounding boxes, and some legend items were paired incorrectly. These errors highlight that the primary limitation of the current approach is layout sensitivity: tightly packed or multi-column legends remain challenging, as the model struggles to distinguish between neighboring entries.

While our evaluation used 40 annotated maps, the dataset covers a range of U.S. Geological Survey (USGS) map styles, including different years, scales, and layout formats.

\section{Conclusion}
\label{sec:conclusion}

In this paper, we presented a method for detecting legend items and their corresponding descriptions on historical geological maps using GPT-4o in an in-context learning (ICL) setting. By combining a layout-aware segmentation step with structured JSON prompting, our approach produces machine-readable bounding boxes without additional training or large-scale annotations. Experiments on USGS maps demonstrated that the method outperforms baseline approaches, and accuracy improves as more in-context examples are provided. The structured outputs integrate naturally with larger digitization workflows\cite{digmapper}.

Our evaluation also revealed limitations: tightly packed multi-column layouts and irregular legend symbols remain challenging, and the model sometimes produces oversized bounding boxes. These findings suggest that prompt design and example selection are crucial factors in maximizing performance.

\begin{acks}
This work is based upon works supported in part by the Defense Advanced Research Projects Agency (DARPA) under Agreement No. HR00112390132 and Contract No. 140D0423C0093.
\end{acks}

\bibliographystyle{abbrv}
\bibliography{sigspatialgeosearch2025}

\begin{thebibliography}{10}

\bibitem{arteaga2013historical}
M.~G. Arteaga.
\newblock Historical map polygon and feature extractor.
\newblock In {\em Proceedings of the 1st ACM SIGSPATIAL International Workshop on MapInteraction}, pages 66--71, 2013.

\bibitem{budig2015active}
B.~Budig and T.~C. van Dijk.
\newblock Active learning for classifying template matches in historical maps.
\newblock In {\em Discovery Science: 18th International Conference, DS 2015, Banff, AB, Canada, October 4-6, 2015. Proceedings 18}, pages 33--47. Springer, 2015.

\bibitem{chiang2009automatic}
Y.-Y. Chiang, C.~A. Knoblock, C.~Shahabi, and C.-C. Chen.
\newblock Automatic and accurate extraction of road intersections from raster maps.
\newblock {\em GeoInformatica}, 13:121--157, 2009.

\bibitem{incontext_survey}
Q.~Dong, L.~Li, D.~Dai, C.~Zheng, J.~Ma, R.~Li, H.~Xia, J.~Xu, Z.~Wu, T.~Liu, et~al.
\newblock A survey on in-context learning.
\newblock {\em arXiv preprint arXiv:2301.00234}, 2022.

\bibitem{digmapper}
W.~Duan, M.~Gerlek, S.~Minton, C.~Knoblock, F.~Lin, T.~Chen, L.~Jang, S.~Kirsanova, Z.~Li, Y.~Lin, and Y.-Y. Chiang.
\newblock Digmapper: A modular system for automated geologic map digitization.
\newblock 06 2025.

\bibitem{gede2021automatic}
M.~Gede and L.~Varga.
\newblock Automatic georeferencing of topographic map sheets using opencv and tesseract.
\newblock In {\em Proceedings of the ICA}, volume~4, pages 1--4. Copernicus GmbH, 2021.

\bibitem{darpa_usgs_2025_dataset}
{Goldman, M. A. and Rosera, J. M. and Lederer, G. W. and Graham, G. E. and Mishra, A. and Yepremyan, A.}
\newblock {Training and validation data from the AI for Critical Mineral Assessment Competition (ver. 2.0, July 2025)}.
\newblock U.S. Geological Survey data release, 2024.
\newblock DOI: 10.5066/P9FXSPT1.

\bibitem{heitzler2020cartographic}
M.~Heitzler and L.~Hurni.
\newblock Cartographic reconstruction of building footprints from historical maps: A study on the swiss siegfried map.
\newblock {\em Transactions in GIS}, 24(2):442--461, 2020.

\bibitem{howe2019deformable}
N.~R. Howe, J.~Weinman, J.~Gouwar, and A.~Shamji.
\newblock Deformable part models for automatically georeferencing historical map images.
\newblock In {\em Proceedings of the 27th ACM SIGSPATIAL International Conference on Advances in Geographic Information Systems}, pages 540--543, 2019.

\bibitem{huang2023pointsymbol}
W.~Huang, Q.~Sun, A.~Yu, W.~Guo, Q.~Xu, B.~Wen, and L.~Xu.
\newblock Leveraging deep convolutional neural network for point symbol recognition in scanned topographic maps.
\newblock {\em ISPRS International Journal of Geo-Information}, 12(3):128, 2023.

\bibitem{layoutlmv3}
Y.~Huang, T.~Lv, L.~Cui, Y.~Lu, and F.~Wei.
\newblock Layoutlmv3: Pre-training for document ai with unified text and image masking.
\newblock In {\em Proceedings of the 30th ACM international conference on multimedia}, pages 4083--4091, 2022.

\bibitem{kim2023mapkurator}
J.~Kim, Z.~Li, Y.~Lin, M.~Namgung, L.~Jang, and Y.-Y. Chiang.
\newblock The mapkurator system: A complete pipeline for extracting and linking text from historical maps.
\newblock In {\em Proceedings of the 31st ACM International Conference on Advances in Geographic Information Systems}, SIGSPATIAL '23, New York, NY, USA, 2023. Association for Computing Machinery.

\bibitem{li2020automatic}
Z.~Li, Y.-Y. Chiang, S.~Tavakkol, B.~Shbita, J.~H. Uhl, S.~Leyk, and C.~A. Knoblock.
\newblock An automatic approach for generating rich, linked geo-metadata from historical map images.
\newblock In {\em Proceedings of the 26th ACM SIGKDD International Conference on Knowledge Discovery \& Data Mining}, pages 3290--3298, 2020.

\bibitem{li2023geolm}
Z.~Li, W.~Zhou, Y.-Y. Chiang, and M.~Chen.
\newblock Geolm: Empowering language models for geospatially grounded language understanding.
\newblock {\em arXiv preprint arXiv:2310.14478}, 2023.

\bibitem{lin2024hyper}
Y.~Lin and Y.-Y. Chiang.
\newblock Hyper-local deformable transformers for text spotting on historical maps.
\newblock In {\em Proceedings of the 30th ACM SIGKDD Conference on Knowledge Discovery and Data Mining}, pages 5387--5397, 2024.

\bibitem{luft2021automatic}
J.~Luft and J.~Schiewe.
\newblock Automatic content-based georeferencing of historical topographic maps.
\newblock {\em Transactions in GIS}, 25(6):2888--2906, 2021.

\bibitem{miao2017symbols}
Q.~Miao, P.~Xu, X.~Li, J.~Song, W.~Li, and Y.~Yang.
\newblock The recognition of the point symbols in the scanned topographic maps.
\newblock {\em IEEE Transactions on Image Processing}, 26(6):2751--2766, 2017.

\bibitem{gpt4o}
OpenAI.
\newblock Gpt-4o technical report.
\newblock \url{https://openai.com/research/gpt-4o}, 2024.
\newblock Accessed: 2025-08-27.

\bibitem{layoutparser}
Z.~Shen, R.~Zhang, M.~Dell, B.~C.~G. Lee, J.~Carlson, and W.~Li.
\newblock Layoutparser: A unified toolkit for deep learning based document image analysis.
\newblock In {\em Document Analysis and Recognition--ICDAR 2021: 16th International Conference, Lausanne, Switzerland, September 5--10, 2021, Proceedings, Part I 16}, pages 131--146. Springer, 2021.

\bibitem{wu2023cross}
S.~Wu, Y.~Chen, K.~Schindler, and L.~Hurni.
\newblock Cross-attention spatio-temporal context transformer for semantic segmentation of historical maps.
\newblock In {\em Proceedings of the 31st ACM International Conference on Advances in Geographic Information Systems}, pages 1--9, 2023.

\end{thebibliography}

\end{document}